\pdfoutput=1

\documentclass[11pt]{article}

\usepackage{acl}

\usepackage{times}
\usepackage{latexsym}
\usepackage{subcaption}
\usepackage[T1]{fontenc}

\usepackage[utf8]{inputenc}

\usepackage{microtype}
\usepackage{graphicx}
\usepackage{hyperref}       
\usepackage{url}            
\usepackage{booktabs}       
\usepackage{amsfonts}       
\usepackage{nicefrac}       
\usepackage{microtype}      
\usepackage{xcolor}         
\usepackage{tcolorbox}
\usepackage{amsthm,enumitem}
\usepackage{amssymb}
\usepackage{amsmath}
\usepackage{listings}
\usepackage{mathrsfs}
\usepackage{subcaption}
\usepackage{soul}
\usepackage{cleveref}
\sethlcolor{yellow}

\DeclareRobustCommand{\parhead}[1]{\textbf{#1}~}
\newcommand{\E}[2]{\mathbb{E}_{#1}\left[#2\right]}
\newtheorem{theorem}{Theorem}

\usepackage[algoruled,algo2e]{algorithm2e}
\usepackage{listings}
\usepackage{fancyvrb}
\fvset{fontsize=\normalsize}
\usepackage{algorithm}
\usepackage{algorithmic}

\crefname{equation}{eq.}{eqs.}
\Crefname{equation}{Eq.}{Eqs.}
\crefname{theorem}{thm.}{thms.}
\Crefname{theorem}{Thm.}{Thms.}
\crefname{figure}{fig.}{figs.}
\Crefname{figure}{Fig.}{Figs.}

\title{Causal Inference for Human-Language Model Collaboration}

\author{Bohan Zhang \\
  University of Michigan \\
  Ann Arbor, MI, USA\\
  \texttt{zbohan@umich.edu} \\\And
  Yixin Wang \\
  University of Michigan\\
   Ann Arbor, MI, USA\\
  \texttt{yixinw@umich.edu} \\ \And
 Paramveer S. Dhillon \\
 University of Michigan\\
  Ann Arbor, MI, USA\\
  \texttt{dhillonp@umich.edu} \\}

\begin{document}
\maketitle


\begin{abstract}
In this paper, we examine the collaborative dynamics between humans
and language models (LMs), where the interactions typically involve
LMs proposing text segments and humans editing or responding to these
proposals. Productive engagement with LMs in such scenarios necessitates that humans discern effective text-based interaction strategies, such as editing and response styles, from historical human-LM interactions. This objective is inherently causal, driven by the counterfactual `what-if' question: how would the outcome of collaboration change if humans employed a different text editing/refinement strategy? A key challenge in answering this causal inference question is formulating an appropriate causal estimand: the conventional average treatment effect (ATE) estimand is inapplicable to text-based treatments due to their high dimensionality. To address this concern, we introduce a new causal estimand\textemdash {\it Incremental Stylistic Effect (ISE)}, which characterizes the average impact of infinitesimally shifting a text towards a specific style, such as increasing formality. We establish the conditions for the non-parametric identification of ISE. Building on this, we develop {\it CausalCollab}, an algorithm designed to estimate the ISE of various interaction strategies in dynamic human-LM collaborations. Our empirical investigations across three distinct human-LM collaboration scenarios reveal that {\it CausalCollab} effectively reduces confounding and significantly improves counterfactual estimation over a set of competitive baselines.
\end{abstract}

\section{Introduction}
\begin{figure*}
    \centering
    \includegraphics[width=0.9\textwidth]{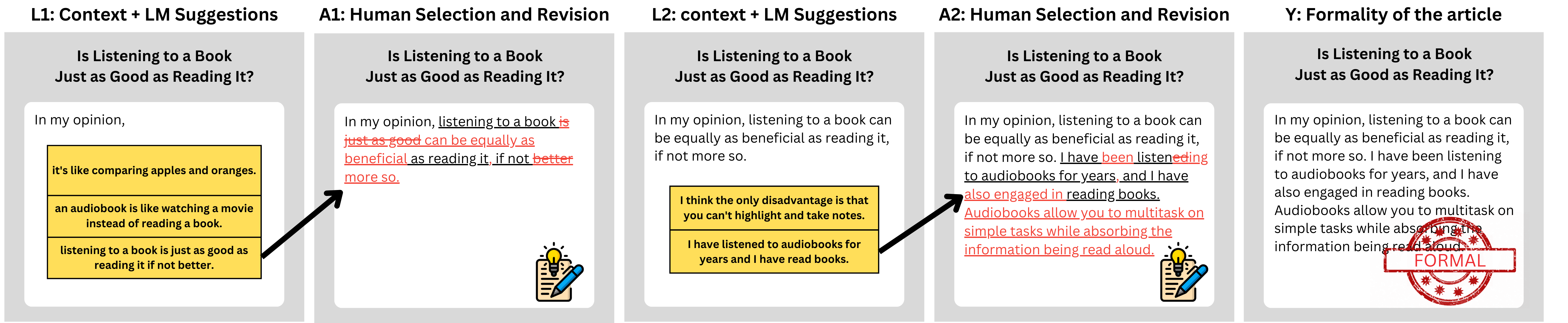}
    \caption{An interactive view of human-LM Collaborative Writing. The writer iteratively selected and rewrote suggestions from the LM to make the article have a better outcome.}
    \label{fig:descriptive-figure}
\end{figure*}

Dialog agents like ChatGPT and Claude, built on Pretrained Language Models (LMs), have significantly transformed the landscape of text generation, showcasing extraordinary performance improvements across a wide range of tasks~\cite{brown2020language, openai2023gpt4,chowdhery2022palm}. This shift has heralded a new era in interaction design, particularly in dialog systems. As illustrated in Figure \ref{fig:descriptive-figure}, these systems are designed for interactive human collaboration, involving sequential actions such as editing, rephrasing, or revising text~\cite{10.1145/3491102.3502030}. This innovative design has empowered users to augment their own capabilities in various fields, including data analysis, customer support, and social media strategy formulation~\cite{brynjolfsson2023generative,noy2023experimental,epstein2023art}. Consequently, it has become crucial for users to learn how to adeptly collaborate with these LMs to fully harness their potential. {\it This paper investigates how users can optimize their collaboration with LMs by drawing on insights from historical human-LM interactions.} We primarily seek to identify effective collaboration strategies from past dialogues, with the goal of enhancing the synergy between human intuition and machine intelligence in these sophisticated dialog systems.

Improving human-LM collaboration by having humans learn from past human-LM interactions is an inherently causal problem. Applying editing strategies from past successful collaborations may not always be effective, since the success of these strategies could be confounded by specific prompt setups.
For instance, editing strategies such as {\it ``increasing the level of politeness in the generated text''} may prove beneficial when collaborating on customer support responses, as it helps to build rapport and maintain a positive relationship with the customer. However, the same strategy may not be as effective when working on a scientific research paper, where a neutral and objective tone is often preferred to convey the findings accurately. Similarly, {\it ``adjusting the text to use more confident language''} may be advantageous when crafting persuasive arguments in an editorial piece, but it could be less appropriate when generating content for a balanced news article, where impartiality is key.

These examples illustrate how editing strategies (increasing politeness and adjusting language confidence) can have varying levels of effectiveness depending on the specific collaboration context (customer support, scientific writing, editorials, and news articles). They underscore the importance of understanding the causal impact of such strategies across different situations to determine their overall usefulness in enhancing human-LM collaboration and bring us to the key counterfactual question:  {\bf How would the collaboration outcome change if we implemented an alternative text editing strategy?}\footnote{We use
the terms ``editing'' and ``refinement'' interchangeably, and they mean the same in our context.} Answering such questions will provide insights into editing strategies that reliably improve human-LM collaboration versus those that only work in certain situational contexts.

This causal inference problem, however, is challenging since it is unclear what constitutes a suitable causal estimand for human-LM collaboration. The traditional causal estimand, Average Treatment Effect (ATE)~\cite{imbens2015causal}, faces limitations in this context due to the high-dimensional nature of text-based treatments, which violate the ``positivity'' assumption. Positivity requires that all possible word sequences occur with non-zero probability in all editing strategies. This condition is readily met in binary treatment situations, where treatments are categorized as either `treated' or `control.' However, it becomes problematic for text-based treatments. Numerous word sequences either fail to form coherent sentences or are implausible as human edits, resulting in some configurations having a zero probability of occurring. Hence, the question arises: {\bf What is an appropriate causal estimand for human-LM collaboration?} This causal estimand must account for the distinctive characteristics of text as a treatment, particularly its high-dimensional and non-binary nature.

In this paper, we propose a novel causal estimand for human-LM collaboration\textemdash Incremental Stylistic Effect (ISE). ISE is based on the key insight that instead of focusing on the effect of specific edits (e.g., insertion/deletion/rephrasing of words) as effective collaboration (text refinement) strategies, we should instead focus on the implications of these edits on text style. For instance, consider a financial analyst collaborating with an LM to write a quarterly report for clients. Multiple edits could make the writing more formal, e.g., removing contractions (`can't' $\rightarrow$ `cannot') or replacing phrasal verbs with more precise terms. In this case, the ISE of ``formalizing'' text would measure the cumulative incremental effect of systematically enhancing formality through any such edits. If the analyst finds that increasing formality improves client evaluations (outcome), they learn that a more formal writing style is effective and can implement appropriate edits in future reports to increase formality (the exact wording changes don't matter). ISE is also more practical to communicate and is {\it actionable}. Instead of recommending specific edits, which may not always be applicable, it advises users on broader stylistic changes that are likely to enhance collaboration outcomes. Finally, the universal applicability of these stylistic changes meets the positivity condition in causal inference, as there is always the possibility to make any text more formal (in our stylized example).

Next, we present an algorithm {\it CausalCollab}, to evaluate the effectiveness of human-LM collaboration strategies over time, using the Incremental Stylistic Effect (ISE) as its guiding estimand. This algorithm operates by identifying and analyzing prevalent stylistic changes in past human-LM interactions. It then evaluates the impact of these changes in various dynamic human-LM collaboration contexts. Our empirical studies, encompassing three distinct scenarios of human-LM collaboration, demonstrate that {\it CausalCollab} is effective in mitigating confounding factors and enhancing counterfactual estimation and provides valuable insights for humans to improve their collaboration strategies with LMs.

\parhead{Contributions:} This paper makes the following contributions:
\begin{itemize}[leftmargin=*,topsep=0pt, itemsep=0pt]
\setlength{\itemsep}{0pt}     
\setlength{\parskip}{0pt}     
\item Formalizing the problem of dynamic human-LM interaction as a causal inference problem.
\item Introducing a novel causal estimand for human-LM collaboration\textemdash Incremental Stylistic Effect (ISE) which addresses the issues of high-dimensionality for text-based treatments.
\item Providing theory establishing identification conditions for ISE.
\item Proposing a new algorithm called {\it CausalCollab} that employs ISE to effectively extract key editing strategies for human-LM collaborations.
\item Thorough, empirical validation of {\it CausalCollab} on three datasets establishing its superior ability for counterfactual prediction.
\end{itemize}



\section{Related Work:}
Our work is related to two strands of prior work.

\underline{\emph{Human-LM Collaboration.}} Building on the foundational
research in human-LM collaboration across text generation
\cite{chakrabarty-etal-2022-help, goldfarb-tarrant-etal-2019-plan},
dialogue \cite{gabriel2019further, bonaldi-etal-2022-human}, and
summarization \cite{avinesh2018sherlock,
shapira-etal-2021-extending}, our work delves into optimizing human
interaction strategies with LMs. We focus on understanding the causal
impacts of various editing strategies, such as the editing strategies
employed by systems like R3 \cite{du-etal-2022-read} and CoAuthor
\cite{10.1145/3491102.3502030}. Our key contribution is the
development of the Incremental Stylistic Effect (ISE) estimand (and
the associated {\it CausalCollab} algorithm), which assesses the
cumulative incremental effect of style-based edits on collaboration
outcomes. This novel approach guides how humans can adapt their
collaboration methods with LMs more effectively, enhancing the synergy
between human intuition and machine intelligence in complex tasks like
those explored in TaleBrush~\cite{10.1145/3491102.3501819} and
Dramatron~\cite{mirowski2023co}. By offering a framework for analyzing
and optimizing human-LM interactions, we aim to improve both the
practical application and theoretical understanding of these human-LM
collaboration dynamics.

\underline{\emph{Causal Inference for Text:}}
Our research is also situated within a rapidly evolving landscape of
studies applying causal methods to language tasks, particularly in the
context of human-LM collaborations. Influential works such as
\citet{pmlr-v124-veitch20a} have pioneered the use of text embeddings
from LMs and topic modeling to address textual confounding, improving
treatment effect estimation. This approach is further expanded by a
series of studies~\cite{egami2022make, roberts2020adjusting,
ijcai2019p0259, pryzant-etal-2021-causal}, which propose learning
latent representations of high-dimensional texts as confounders or
treatments, utilizing topic modeling and Variational Autoencoders
(VAEs). These studies highlight the importance of distilling
low-dimensional latent representations for accurate treatment effect
estimation from complex, high-dimensional data such as
text~\cite{louizos2017causal, kim2021counterfactual}.

Our work distinguishes from this line of work in addressing the
dynamic nature of human-LM collaborations and proposing a novel causal
estimand\textemdash Incremental Stylistic Effect (ISE) for this
scenario. Further, unlike prior work, we employ a novel combination of
G-estimation~\cite{taubman2009intervening,
naimi2014constructing,van2011targeted, petersen2012diagnosing} and
Conditional Variational Autoencoders (CVAE) for learning
low-dimensional latent representations in a setup with time-varying
textual treatments.
\section{Causal Inference for Human-LM Collaboration}
We begin with framing human-LM collaboration as a causal inference
problem.



\subsection{A causal perspective on Human-LM collaboration} We
conceptualize the interaction between a human and a language model
(LM) as a sequential series of human actions, each focused on
optimizing task outcomes. In these iterative collaborations, humans
consistently adjust the responses generated by LMs to attain the
desired results. This approach is supported by various studies,
including~\citet{xu2023baize,bonaldi-etal-2022-human,du-etal-2022-understanding-iterative,
lee2023evaluating}, which explore the nuances of human-LM interactions
in achieving specific goals.

To establish our notation, we represent the refinement action (such as the post-refinement text)
performed by user \(i\) at time \(t\) as \(A_{it}\), with
$t~=~1~\ldots~T$. Further, let's denote $Y_{i}(a_{1}, \ldots, a_{T})$
as the outcome rating (like a quality score) of the text, assuming
user $i$ \emph{had executed} the refinement $a_{1}$ at time $t=1$,
$a_{2}$ at time $t=2$, \ldots, and $a_{T}$ at time $t=T$ during their
interactions with the LM. This outcome \(Y_{i}(a_{1}, \ldots, a_{T})\)
is conceptualized as a \emph{potential outcome} or
\emph{counterfactual
outcome}~\citep{imbens2015causal,pearl2009causality,hernan2010causal}.
In this context, the sequence of user refinements is regarded as the
``treatment,'' and the outcome rating of the text is the ``outcome.''
The potential outcome \(Y_{i}(a_{1}, \ldots, a_{T})\) is observable
only when user \(i\) actually performs these refinements, that is,
when \(A_{it} = a_t\) for all \(t=1, \ldots, T\). All other potential
outcomes \(Y_{i}(a'_{1}, \ldots, a'_{T})\), where \(a'_t \ne A_{it}\),
are unobserved.

\begin{figure}[htbp]
    \centering
    \includegraphics[width=0.9\linewidth]{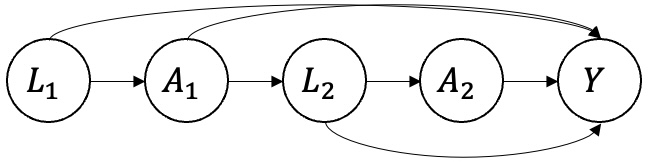}
    \caption{The causal graph of human-LM collaboration ($T$=2).}
    \label{fig:two-step dag}
\end{figure}

At each time step \( t \), the refinement action \( A_{it} \) by user
\( i \) is typically in response to the output of the LM, denoted as
\( L_{it} \). For instance, \( L_{it} \) could be the LM's suggestions
for completing an unfinished essay at time \( t \) during the
interaction with user \( i \), while \( A_{it} \) might involve
selective editing, refinement, and rewriting by the human participant.
This action \( A_{it} \) then acts as a prompt for the LM's subsequent
response \( L_{i, t+1} \) at the following time step. The two-step
process of human-LM collaboration can be visualized as a directed
acyclic graph (DAG), as shown in Figure~\ref{fig:two-step dag}, with the
user subscript \( i \) omitted for simplicity; this DAG can be readily
extended to more than two steps. (A real-world example of our 
human-LM collaborative writing setup is 
shown in Figure~\ref{fig:descriptive-figure}.)

\vspace{-.3cm}


Our goal is to evaluate the impact of various user refinements on the
final text outcome by examining a dataset comprising historical
human-LM interactions. This dataset typically includes: (1) the
sequence of actions performed by previous users, denoted as
\(\{\{A_{it}\}_{i=1}^I\}_{t=1}^T\), (2) the LM's responses at each
time step, represented by \(\{\{L_{it}\}_{i=1}^I\}_{t=1}^T\), and (3)
the observed outcomes, such as the quality scores of the texts,
labeled as \(\{Y_i\}_{i=1}^I\), for all the interactions
$i~=~1~\ldots~I$. These observed outcomes are the potential outcomes
for the actual action taken by the user, expressed as \(Y_i =
Y_i(A_{i1}, \ldots, A_{iT})\).

To evaluate the efficacy of user refinements, we are often interested
in the causal estimand known as the average treatment effect (ATE) of
a given action sequence \( (a_{1}, \ldots, a_{T}) \). This is
mathematically represented as \( \E{}{Y_{i}(a_{1}, \ldots, a_{T}) -
Y_{i}(a^\varnothing_{1}, \ldots, a^\varnothing_{T})} \), where \( a_1,
\ldots, a_T \) denotes the sequence of user actions. This expectation
is calculated across users. The sequence \( a^\varnothing_{1}, \ldots,
a^\varnothing_{T} \) represents a baseline or `null' action sequence,
wherein the user does not perform any refinements.



\subsection{The challenge of textual treatments in causal human-LM
collaboration} The causal estimand of ATE can be identified using
standard tools in dynamic treatment regimes when the treatment is
binary. For example, when each user can only choose between two
actions to execute at each time step, we can identify the ATE using
the $g$-formula,
\begin{align*}
&\E{}{Y_{i}(a_1, \ldots, a_T)} = \int \E{}{Y_i|\bar{A}_{iT}=\bar{a}_T, \bar{L}_{iT}=\bar{l}_T}\\&\times\prod_{t=1}^T P(L_{it}=l_t|\bar{A}_{i,t-1}=\bar{a}_{t-1}, \bar{L}_{i,t-1}=\bar{l}_{t-1})\mathrm{d}\bar{l}_T,
\end{align*}
where $\Bar{A}_{it}=A_{i,1:t}, \bar{a}_{t} = a_{1:t}$,
$\Bar{L}_{it}=L_{i,1:t}$, and $\bar{l}_{t} =
l_{1:t}$~\citep{hernan2010causal}.

This $g$-formula for ATE requires two conditions: the sequential
exchangeability condition and the positivity (a.k.a. overlap)
condition. The sequential exchangeability condition roughly requires
that $\Bar{L}_{it}, \Bar{A}_{i,t-1}$ capture all confounders
(variables that affect both $A_{it}$ and $Y_i$) at time $t$. This
requirement is often satisfied in human-LM collaboration: the LM's
text response $\Bar{L}_{it}$ and the previous actions
$\Bar{A}_{i,t-1}$ are the only factors that affect both the outcome
and the human's actions (cf.~\Cref{fig:two-step dag}).

The second positivity condition, however, is often violated in
handling textual treatments in human-LM collaboration. The positivity
condition loosely requires that all possible values of the actions
$a_t$ shall occur with nonzero probability given any confounder values
$\bar{l}_{t}$. This condition is challenging to satisfy even when
$T=1$; the reason is that each $a_t$ is a textual treatment\textemdash
composed of a sequence of words\textemdash hence high-dimensional. It
often violates the positivity condition in that many values of the
treatment are not possible: a sequence of randomly chosen words likely
occurs with zero probability as a human refinement strategy. This
violation of positivity renders the ATE hard to identify in human-LM
collaboration.

Moreover, the causal estimand involving the average treatment effect
(ATE), represented as \( \E{}{Y_{i}(a_{1}, \ldots, a_{T}) -
Y_{i}(a^\varnothing_{1}, \ldots, a^\varnothing_{T})} \), holds limited
practical significance for enhancing future human-LM interactions.
Specifically, the treatment \( a_t \) usually refers to the
post-refinement human text; for example, setting the text to ``the
prince proposed to the princess.'' The ATE in this context quantifies
the effect of this specific post-refinement text on the outcome,
regardless of the context (like the LM's preceding prompts or human
responses). However, this ATE might not be practically useful since
such text refinement cannot be universally applied across different
human-LM collaboration scenarios. For instance, applying this edit is
illogical when the initial LM prompt pertains to the animal world,
rather than a story about a prince and princess. Consequently, the
traditional causal estimand of ATE is unsuitable for causal inference
in scenarios involving textual treatments.

\subsection{Incremental Stylistic Effect (ISE): A causal estimand for
textual treatments} Considering the limitations of the average
treatment effect (ATE) in the context of textual treatments for
human-LM collaboration, it becomes essential to ask: what is an
appropriate causal estimand for these types of collaborations? We
propose the causal estimand\textemdash\emph{incremental stylistic
effect} (ISE), which focuses on the impact of {\it style change} in
human refinements, a concept we will define more formally later.
Focusing on stylistic modifications helps overcome the challenges
typically encountered with raw textual treatments. {\it Style Change}
as a treatment (e.g., editing the text to be more formal) can often meet the
positivity condition; that is, all variations of the `style change'
treatment have a nonzero probability of occurrence since any style
modification can be applied regardless of the confounder values.
Moreover, it is also more practically relevant than ATE: regardless of
the context, a user can always apply the ``style change'' treatment to
the text, e.g., rewrite to be more formal.

More formally, we define the ``style change'' treatment as an
intervention on some dimension-reducing function $f_t(\cdot)$ of the
post-refinement text and the LM's previous response $f_t(A_{it},
L_{it};
\bar{A}_{i,t-1}, \bar{L}_{i,t-1})$ (abbreviated as $f_t(\bar{A}_{it},
\bar{L}_{it})$) at each time step; it captures how $A_{it}$ is
different from $L_{it}$ given previous histories $\bar{A}_{i,t-1},
\bar{L}_{i,t-1}$, hence revealing the style change performed by user
$i$ at time $t$. Thus we define \emph{incremental stylistic effect}
(ISE)  as
\vspace{-.18cm}
\begin{align}
\mathrm{ISE}=\lim_{\delta\rightarrow
0}[&\E{}{Y_i(\{f_t(\bar{a}_t, \bar{L}_{it})+\delta\}_{t=1}^T)}\nonumber\\ &
-\E{}{Y_i(\{f_t(\bar{a}_t, \bar{L}_{it})\}_{t=1}^T)}]/\delta,
\label{eq:ISE}
\end{align}
where $Y_i(\{f_t(\bar{a}_t, \bar{L}_{it})\}_{t=1}^T)$ denotes the
potential outcome of executing the style change sequence $(f_1,
\ldots, f_T)$; $f_t(\bar{a}_t, \bar{L}_{it})$ extracts the style feature of
interest on text $a_t, L_{it}$ given history up to $t-1$; it can
characterize how $a_t$ differs from $L_{it}$ in terms of politeness
(or formality), for instance. At a high level, ISE characterizes the
impact of an infinitesimal style change sequence on the final outcome;
we focus on infinitesimal changes due to the hardness to quantify
style change.

We next define the conditional expected potential outcome over style change as
follows,
\begin{align}
&\E{}{Y_i(\{f_t(\bar{a}_t, \bar{L}_{it})\}_{t=1}^T)\mid \bar{L}_{iT}}\nonumber\\
& \triangleq \int \{\E{}{Y_i(\bar{a}'_T)\mid \bar{L}_{iT}}\nonumber\\
& \quad \times \prod_{t=1}^T P(A_{it}=\bar{a}'_t\mid \bar{L}_{it},\bar{A}_{i,t-1},\nonumber\\
& \qquad \qquad f_t(\bar{A}_{it}, \bar{L}_{it})=f_t(\bar{a}_t, \bar{L}_{it}))\}\mathrm{d}\bar{a}'_T.
\label{eq:potential-style-outcome}
\end{align}
It describes the conditional potential outcome of the style change sequence
$f_{1:T}$ as the average outcome over all refinements $\bar{a}'_T$
that correspond to the same style change $f_{1:T}(\bar{a}_t,
\bar{L}_{it})$ given contexts $\bar{L}_{it}$'s. This definition aligns
with functional
interventions~\citep{puli2020causal,correa2020calculus,pearl2009causality,wang2021desiderata},
where interventions are performed on some deterministic functions of
high-dimensional treatments. It reflects the goal of assessing the
impact of style change, regardless of context or specific text edit.


To estimate ISE from observational human-LM interaction data, we
establish nonparametric identification conditions for the causal
estimand ISE.

\begin{theorem}[Non-parametric identification of ISE]
\label{thm:id-ise}
Under (1) the positivity condition for $\{f_t(\cdot, \cdot)\}_{t=1}^T$,
and (2) the sequential exchangeability condition, the ISE of the style
change sequence $f_{1:T}$ can be non-parametrically identified by
plugging \Cref{eq:id-style-change} into \Cref{eq:ISE},
\begin{align} &\E{}{Y_i(\{f_t(\bar{a}_t,
\bar{L}_{it})\}_{t=1}^T)}\nonumber \\ &=  \int
\mathbb{E}[Y_i|\{f_s(\bar{A}_{is}, \bar{L}_{is})=f_s(\bar{a}_s,
\bar{L}_{is}) \}_{1}^T, \bar{L}_{iT}=\bar{l}_T]\nonumber\\
&\times\prod_{t=1}^T P(\bar{L}_{it}=\bar{l}_t \mid\{f_s(\bar{A}_{is},
\bar{L}_{is})=f_s(\bar{a}_s, \bar{L}_{is}) \}_{s=1}^{t-1}, \nonumber\\
&\qquad\qquad\bar{L}_{i,t-1}=\bar{l}_{t-1})\,\mathrm{d}\bar{l}_T.
\label{eq:id-style-change}
\end{align}
\end{theorem}

The proof of \Cref{thm:id-ise} is in \Cref{sec:id-ise-proof}.
\Cref{thm:id-ise} generalizes the classical $g$-formula for ATE: when
$f_t(\cdot)$'s are identity functions as opposed to
dimensionality-reducing functions, \Cref{eq:id-style-change} recovers
the classical $g$-formula~\citep{hernan2010causal}. The sequential
exchangeability condition requires $Y_i(\bar{a}_t) \perp A_{it} |
\Bar{A}_{i,t-1}, \Bar{L}_t, \forall
t=1,\ldots,T$~\citep{hernan2010causal}. The positivity condition
requires that $P(f_t(A_{it}, L_{it}) \in \mathcal{V}| \bar{L}_{i,t-1})
> 0$ for any set $\mathcal{V}$ that satisfies $P(f_t(A_{it}, L_{it})
\in
\mathcal{V}) > 0$~\citep{imbens2015causal}. \Cref{thm:id-ise} shows
that, if we are interested in some style changes $f_t$ that are common
in historical human-LM interactions, one can employ
\Cref{eq:id-style-change,eq:ISE} to estimate their ISE. Such common
style changes are more likely to satisfy the positivity condition;
loosely, these style changes can be applied to any text, regardless of
context.


\subsection{\emph{CausalCollab}: An algorithm for dynamic human-LM Collaboration} 

We next operationalize \Cref{thm:id-ise} to perform causal inference
for human-LM collaboration.

The first step is to extract common style changes $f_{1:T}$ in historical human-LM collaboration dataset. These are the style changes often adopted by human users; their ISE can be calculated using \Cref{thm:id-ise}. We perform this extraction by fitting a conditional variational autoencoder
(CVAE)~\citep{zhao2017learning,lopez2017conditional,mishra2018generative,kim2021conditional}
to all historical human-LM interactions $\{\Bar{A}_{iT},
\Bar{L}_{iT}\}_{i=1}^I$:
\begin{align}
\label{eq:CVAE-model}
z_{it} &\sim \mathcal{N}(0, \sigma^2I_K),\\
A_{it}\mid \Bar{A}_{i,t-1}, z_{it}, \Bar{L}_{it}&\sim \mathcal{N}(h(\Bar{A}_{i,t-1}, z_{it}, \Bar{L}_{it}), \sigma^2I_d),\nonumber
\end{align}
where $K \ll d$. We use variational
inference~\citep{blei2017variational} to infer the latent variables
$z_{it}$, employing a variational approximating family of $z_{it}~\sim~
\mathcal{N}(\hat{f}_t(\Bar{A}_{it}, \Bar{L}_{it}), \sigma^2)$.
Specifically, we maximize the evidence lower bound (ELBO) of the CVAE over $h$ and $\hat{f}$~\citep{kingma2013auto}. The resulting posterior mean of the latents $z_{it}$ reveals the common style changes in the
dataset. One may choose the style changes of interest as the posterior
mean of $z_{it}$ in the CVAE fit, i.e. setting
$f_t=\hat{f}_t$.\footnote{This procedure recovers conditional
probabilistic PCA, if $h(\cdot)$ is constrained to be
linear~\citep{tipping1999probabilistic}.}

Given the chosen style changes, we use \Cref{thm:id-ise} to estimate
their ISEs. To employ \Cref{eq:id-style-change}, we fit an outcome
model of $Y_i$ against $\{\hat{f}_s(A_{is}, L_{is}), \bar{L}_{is},
\bar{A}_{i,s-1}\}$. One example is generalized additive
models\footnote{While any choice of outcome model is valid, we find
that generalized additive models can often help reduce variance in the
later Monte Carlo estimation~\citep{kroese2013handbook}.}:
$Y_i = b_1(\hat{f}_s(\bar{A}_{is}, \bar{L}_{is})) +
b_2(\bar{L}_{is}) + b_3(\bar{A}_{i,s-1}) + \epsilon_i,$
where both $b_{k}(\cdot), k=1,2,3$ are neural networks and
$\epsilon_i\sim \mathcal{N}(0, \sigma^2)$. We then employ Monte Carlo
estimation~\citep{shapiro2003monte} for the integrals in
\Cref{eq:id-style-change,eq:ISE}.  We leave the details of these steps
to \Cref{sec:alg-detail}.

These two steps constitute \emph{CausalCollab}
(\Cref{alg:causalcollab}), an algorithm that performs causal inference
for human-LM collaboration.


\begin{algorithm}[t]
  \DontPrintSemicolon
  \;
  \KwIn{A dataset of historical human-LM interactions $\{(\bar{A}_{iT}, \bar{L}_{iT}, Y_i)\}_{i=1}^I$.}

  \BlankLine

  \KwOut{The incremental stylistic effect of common style changes $\textsc{ISE}_{f_{1:T}}$.}

  \BlankLine

  \begin{flushleft}
  1. Fit CVAE to the historical data from \Cref{eq:CVAE-model}, and extract the fitted $\hat{f}_{1:T}$ as common style changes;

  \BlankLine

  2. Fit an outcome model for $\{Y_i\}_{i=1}^I$ against $\{\{\hat{f}_s(\bar{A}_{is}, \bar{L}_{is})\}_{s=1}^T\}_{i=1}^I, \bar{L}_{iT}$;
  

  \BlankLine

  3. Estimate the ISE of $\hat{f}_{1:T}$ using \Cref{eq:ISE,eq:id-style-change}.
  \end{flushleft}
  \caption{The \emph{CausalCollab} algorithm}
  \label{alg:causalcollab}
\end{algorithm}


\section{Empirical Studies}

\subsection{Experiment setup and evaluation metrics}

\label{sec:general-data}

\parhead{Evaluation metrics of \emph{CausalCollab}:} The quality of
ISE estimates from \emph{CausalCollab} can be evaluated by evaluating the quality of the intermediate potential
outcome estimates for $\E{}{\{Y_i(\{f_t(a_t, L_{it})\}_{i=1}^T})$. It
is because ISE is the limiting difference between two such potential
outcome estimates. The potential outcome estimates can only be
evaluated using semi-synthetic studies since potential outcomes are in
general not observable in real data~\citep{imbens2015causal}. Thus, we
simulate semi-synthetic data from real human-LM interactions and
assess the closeness between the actual potential outcome
$Y_i(\{f_t(a_t, L_{it})\}_{i=1}^T)$ and the \emph{CausalCollab} estimates of
$\E{}{Y_i(\{f_t(a_t, L_{it})\}_{i=1}^T)}$, uniformly averaging over
all possible values of $a_t$.

\parhead{Confounder, Outcome, Observational and Counterfactual Data:}
To effectively test causal methods, both observational and
counterfactual data are
necessary~\citep{pearl2009causality,imbens2015causal}. Constructing
counterfactual language datasets presents a significant challenge
because generating plausible ``what if'' scenarios requires a deep
understanding of text and is often subjective as the possible
counterfactual space of language is vast. Hence, we use ChatGPT to
generate the counterfactual data thanks to their ability to generate a
large volume of coherent texts given counterfactual
instructions~\cite{li2023large,fryer-etal-2022-flexible}. Thus, if
$Y(\Bar{a}_2)=1$ for an observation, we ask ChatGPT to rewrite
$\Bar{a}_2$ to counterfactual $\Bar{a}_2'$ so that
$Y(\Bar{a}_2')=0$ and vice versa for $Y(\Bar{a}_2)=0$.

We then identify a potential confounding signal $X$ embedded within
the high dimensional $L_i$ that will decide both the outcome $Y$ and
the treatment $\Bar{A}_2$. For example, if the outcome is the
formality of articles in the human-LM collaborative writing task, then
the type of articles (e.g., argumentative or creative) can confound
the impact of human refinement on the outcome. For instance, if, in the
observational dataset, argumentative articles tend to be more formal
and creative ones less formal, then for effectively testing our
proposed ISE estimand, it's essential to see a non-existent or
reversed correlation in the counterfactual data. In other words, we
would need the creative articles to be more formal and the
argumentative articles to be less formal in the counterfactual
scenario.

After identifying the confounding signal, we establish an \(\alpha\sim
\text{split}\) in the dataset to generate the observational and
counterfactual data quantitatively. In the observational data, the
probability \(P(Y=0|X=1)\) is set to \(\alpha\), and \(P(Y=0|X=0)\) to
\(1-\alpha\). Conversely, in the counterfactual data, these
probabilities are reversed: \(P(Y=0|X=1)\) becomes \(1-\alpha\) and
\(P(Y=0|X=0)\) is \(\alpha\). For instance, with \(\alpha=0.2\), in
observational data, an argumentative article (\(X=1\)) has a 80\% (0.8)
chance of being formal (\(Y=1\)), but in counterfactual data, this
likelihood drops to 20\%. Here, \(\alpha\) indicates the strength of
the confounding correlation, with a lower \(\alpha\) (for
\(\alpha<0.5\)) suggesting a stronger correlation. Our data generation 
approach remains the same for different \(\alpha\).

Labeling subjective textual outcomes (Y), like formality, is challenging.
LMs trained on diverse textual datasets, often provide more consistent
and accurate annotations for subjective tasks (such as political
affiliation labeling, relevance assessment, and stance detection) than
humans. Here again we employ ChatGPT for labeling task-specific outcomes.
Detailed information about the prompts used for counterfactual
generation and outcome labeling across different datasets is in
\Cref{sec:promp}.

We employ three human-LM dynamic interaction datasets for our empirical study, 
each focusing on distinct tasks: CoAuthor~\cite{10.1145/3491102.3502030}, 
Baize~\cite{xu2023baize}, and DIALOCONAN~\cite{bonaldi-etal-2022-human}.
CoAuthor features 1445 human-GPT-3 collaborative writing tasks, 
focusing on the dynamic between human edits and language model
suggestions in both creative and argumentative writing.
Baize, created through self-chat with ChatGPT, contains over 
200k dialogues from sources like Quora and StackOverflow, 
focusing on the helpfulness of revised answers in chats. 
DIALOCONAN addresses multi-turn counter-narrative generation 
against hate speech using a hybrid human-machine approach, 
editing dialogues generated by DialoGPT~\cite{zhang-etal-2020-dialogpt} and T5~\cite{raffel2020exploring}. 
The dataset details are in \Cref{sec:data-detail}
and are summarized in \Cref{tab:dataset-sum}.
More implementation details of CVAE and the Monte Carlo estimations  for \emph{CausalCollab} are provided in
\Cref{sec:implementation-detail}. 
For comparison, other than CVAE, we also implement a PCA to 
extract style changes adopted by humans. 
We refer to CVAE and PCA as {\it treatment embeddings} in the results 
discussion as they both attempt to learn low-dimensional embeddings for human treatments.

\begin{table*}
\centering
\resizebox{\textwidth}{!}{%
\begin{tabular}{l|lll}
\toprule
Dataset        & Coauthor~\cite{10.1145/3491102.3502030}             & Baize~\cite{xu2023baize}          & DIALOCONAN~\cite{bonaldi-etal-2022-human}          \\
\midrule
Tasks& human-LM Collaborative Writing & Multi-turn Chat     & Hate Speech Countering  \\
Outcome    & Formality            & Helpfulness     & Effectiveness       \\
Confounding Signal & Type of Text (collaborative/argumentative)        & Confidence (high/low)    & Formality (high/low)  \\
\bottomrule
\end{tabular}
}
\caption{Summary of Datasets}
\label{tab:dataset-sum}
\end{table*}

\begin{table*}[ht]
\centering
\resizebox{\textwidth}{!}{%
\begin{tabular}{l|cc|cc|cc}
\toprule
Dataset & \multicolumn{2}{c|}{Coauthor} & \multicolumn{2}{c|}{Baize} & \multicolumn{2}{c}{DIALOCONAN} \\
Model & Observational & Counterfactual & Observational & Counterfactual & Observational & Counterfactual \\
\midrule
No Adjustment & \large{.188} \small{(.004)} & \large{.353} \small{(.018)} & \large{.270} \small{(.011)} & \large{.351} \small{(.003)} & \large{.287} \small{(.018)}& \large{.489} \small{(.017)}\\
No Adjustment + PCA & \large{.163} \small{(.006)}& \large{.383} \small{(.015)}& \large{.215} \small{(.010)}& \large{.276} \small{(.005)}& \large{.227} \small{(.005)}& \large{.363} \small{(.003)}\\
No Adjustment + CVAE & \large{.173} \small{(.003)} & \large{.407} \small{(.016)} & \large{.218} \small{(.004)}& \large{.276} \small{(.006)}& \large{.236} \small{(.004)}& \large{.357} \small{(.003)}\\
G-E & \large{.213} \small{(.010)} & \large{.252} \small{(.009)}& \large{.266} \small{(.009)}& \large{.346} \small{(.006)}& \large{.283} \small{(.016)}& \large{.488} \small{(.018)}\\
G-E + PCA & \large{.201} \small{(.003)}& \textbf{\large{.219}} \small{(.004)}& \large{.199} \small{(.005)}& \textbf{\large{.232}} \small{(.003)}& \large{.222} \small{(.002)}& \large{.273} \small{(.004)}\\
G-E + CVAE & \large{.216} \small{(.001)} & \textbf{\large{.219}} \small{(.004)} & \large{.202} \small{(.004)}& \textbf{\large{.232}} \small{(.001)}& \large{.227} \small{(.000)}& \textbf{\large{.272}} \small{(.002)} \\
\bottomrule
\end{tabular}
}
\caption{Performances (MSE) of different methods under three datasets. All datasets are $\alpha=0.2\sim \text{split}$ and the noise level $\sigma=1.0$. G-E stands for G-estimation. The numbers are the average of three different random seeds with the standard deviation recorded within the parentheses.}
\label{tab:results}
\end{table*}
\vspace{-.2cm}
\subsection{Results}
\vspace{-.2cm}
\parhead{Quantitative analysis:} The results of using each method for
predicting the outcome in each of the three datasets (averaged over three random seeds) are shown in Table~\ref{tab:results}. The performance is
measured using Mean Squared Error (MSE) so a lower number indicates
better performance. Across all three datasets, G-estimation + CVAE or
PCA (Row 5 and 6) intervention significantly narrows the gap between observational and counterfactual
MSE. Our approach improves counterfactual performance significantly
while maintaining competitive or better observational
performance in all three datasets compared to methods without
adjustments.

The large performance gap between the counterfactual and observational
performance when there is no confounder adjustment shows that baseline
models heavily rely on confounders for prediction. However, using
G-estimation alone without treatment embeddings for adjustment is not
effective in closing the gap for the Baize and DIALOCONAN datasets and
it also performs worse than the one with treatment embeddings in the
CoAuthor dataset. Using treatment embeddings alone also shows
significantly worse performance compared to when it is combined with
G-estimation across all datasets. This suggests that integrating both
G-estimation and treatment embeddings are crucial for causal
observational and counterfactual predictions. CVAE consistently
matches the performance of PCA, which implies that the human strategies 
underlying $A_i$ may be invariant. If a human rewrites 
the LM output to be more formal, they may be rewriting phrases 
from the informal subspaces to formal subspaces, which can be
linear and independent of the words themselves.

\begin{figure}[h]
\vspace{-.2cm}
    \centering
    \begin{subfigure}{.9\linewidth}
        \centering
        \includegraphics[width=\linewidth]{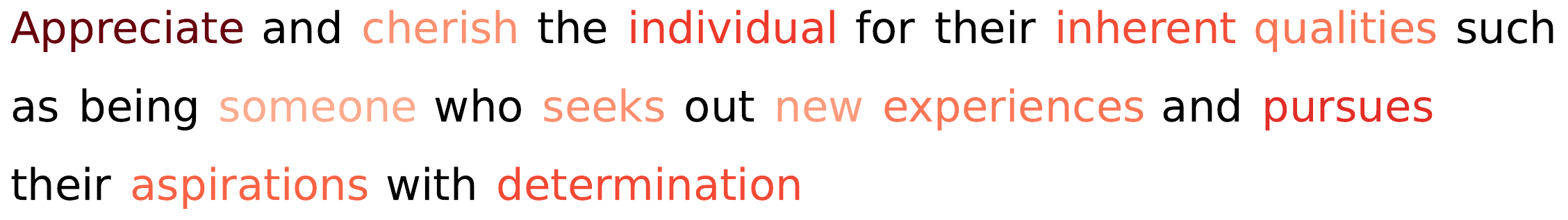}
        \caption{Example 1: short length $A_2$}
        \label{fig:sub1}
    \end{subfigure}%
    \hfill
    \begin{subfigure}{.9\linewidth}
        \centering
        \includegraphics[width=\linewidth]{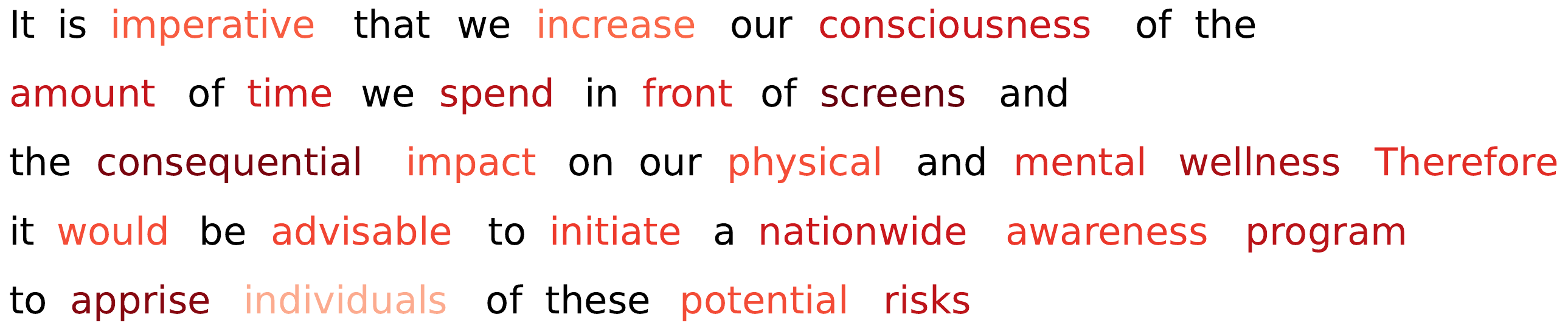}
        \caption{Example 2: middle length $A_2$}
        \label{fig:sub2}
    \end{subfigure}%
    \hfill
    \begin{subfigure}{.9\linewidth}
        \centering
        \includegraphics[width=\linewidth]{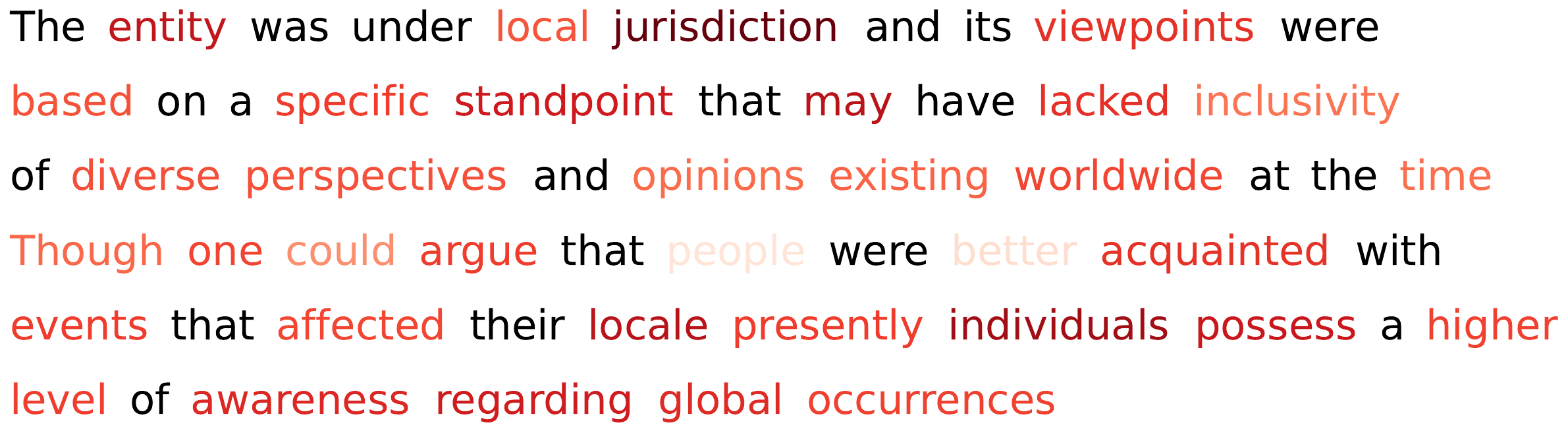}
        \caption{Example 3: long length $A_2$}
        \label{fig:sub3}
    \end{subfigure}
    \caption{Three examples of $A_2$ with different lengths. The color of words is decided by their cosine distance to the treatment $z_2$ learned by the CVAE. The darker the color, the closer a word is to $z_i$. Stopwords are black.}
    \label{fig:qual-example}
\end{figure}

\parhead{Qualitative Analysis: What does $z$ capture?} Our quantitative results show that the learned treatment embeddings successfully narrow the
performance gap between the counterfactual and observational data. But
what does $z$ capture that can help predict the outcomes? We identify
words in the human refinement $A_i$ that are semantically closest to
their learned treatment embeddings $z_i$ from CVAE. As the outcome of
the CoAuthor dataset, \textit{formality}, is more intuitive for
analysis than the other two, our following investigation is based on
the CoAuthor dataset and specifically on the second step of the
collaboration ($A_2$).

\begin{figure*}[htbp]
    \centering
    \includegraphics[width=\textwidth]{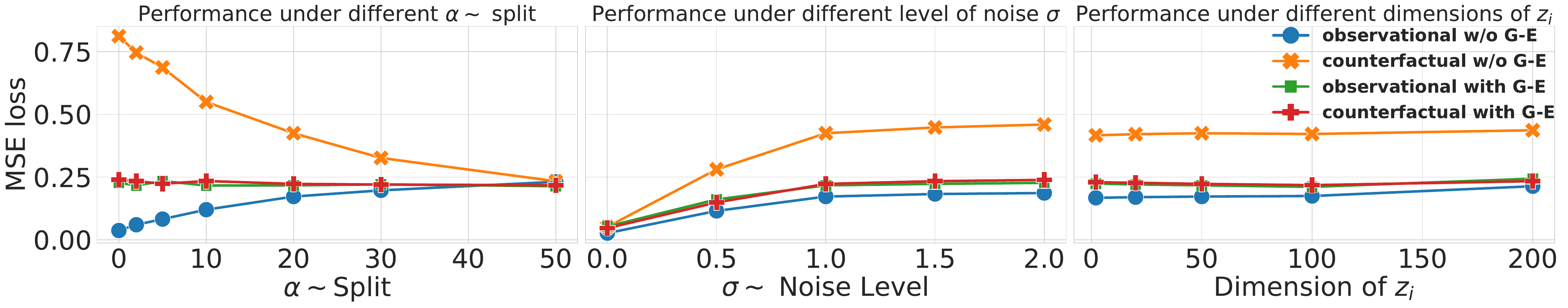}
    \caption{Performances of our methods under different $\alpha \sim$ split, levels of noise $\sigma$, and dimensions of $z_i$ on the coauthor dataset. As $\alpha$ increases, the confounding correlation weakens, and our adjustment maintains similar counterfactual and observational performances. The method is robust to varying noise levels $\sigma$, keeping both counterfactual and observational performance low. The choice of $z_i$ dimension has minimal impact on performance, indicating that predictive treatments can be effectively represented in a low-dimensional space.}
    \label{fig:coauthor_results}
\end{figure*}
To embed words in $A_2$, we trained a 50-dimensional Word2Vec 
model~\citep{mikolov2013efficient} for consistency with $z_2$'s dimension. 
Cosine distances between Word2Vec embeddings of each word in $A_2$ 
and the corresponding $z_2$ were calculated. 
These distances, standardized and represented by color intensity
(darker hues indicating closer proximity to $z_2$ and stop words in black), 
are visualized for three different lengths of $A_2$s in Figure \ref{fig:qual-example}.
The top 3 words that were closest to $z_2$ are: 
`Appreciate', `pursues', and `individual' for Figure \ref{fig:sub1};
`apprise', `consequential', and `wellness' for Figure \ref{fig:sub2};
`jurisdiction', `individuals', `may' for Figure \ref{fig:sub3}.
Conversely, words deemed informal or neutral, such as `someone', `people', and
`better', were observed to be more distant from $z_2$.
The qualitative
findings suggest that the CVAE model is capable of learning
explainable human strategies according to the outcomes of the task.


\parhead{Robustness to 1) $\alpha \sim$ split, 2) noise, and 3) dimension of $z_i$:} 
We assess the robustness of our methods to $\alpha\sim$
split, noise, and the varying dimensions of the CVAE's latent space 
by conducting empirical studies on the CoAuthor dataset. 
These tests were performed with and without G-estimation, 
incorporating $\alpha \in \{0, 2, 5, 10, 20, 30, 50\}$,
noise levels \(\sigma \in \{0, 0.5, 1.0, 1.5, 2.0\}\) 
and dimensions for \(z_i\) set at \(\{2, 20, 50, 100, 200\}\).

As shown in Figure \ref{fig:coauthor_results}, our
adjustment keeps the counterfactual and observational performances
close consistently regardless of the intensity of the confounding
correlations (Note the confounding correlation is stronger as
for smaller $\alpha$, i.e., $\alpha<0.5$.)
When $\alpha$ is $0.5$, the confounder $L_i$ is independent
of the independent variable $A_i$, so the performances on both data
are expected to be close. For noise and dimension of $z_i$, detailed results are provided in the Appendix \ref{app:z_dim-noise}; we observed that our approach significantly outperformed others in scenarios with higher noise levels. Additionally, it demonstrated minimal performance variation across the different dimensions of \(z_i\).


\vspace{-.3cm}
\section{Conclusion}
\vspace{-.3cm}
In conclusion, our work introduces the Incremental Stylistic Effect (ISE) as a novel causal estimand for human-LM collaboration, addressing the challenges posed by high-dimensional text-based treatments in causal inference. We establish the conditions for the non-parametric identification of ISE, providing a solid theoretical foundation for our approach. Building upon this, we propose the {\it CausalCollab} algorithm, which leverages G-estimation for time-varying text variables and Conditional Variational Autoencoders (CVAE) to extract human strategies, enabling the estimation of ISE in practice.

Our research makes significant contributions to the field of human-LM collaboration by offering a fresh perspective on understanding the dynamics of these interactions through a causal lens. By focusing on the stylistic implications of text edits rather than specific wording changes, ISE provides a more practical and actionable approach to identifying effective collaboration strategies. The universal applicability of stylistic changes also ensures that the positivity condition in causal inference is met, overcoming the limitations of traditional causal estimands like Average Treatment Effect (ATE).

We demonstrate the effectiveness of our approach through extensive empirical studies on three diverse human-LM collaboration datasets. The results validate that {\it CausalCollab} significantly narrows the gap between observational and counterfactual evaluations, highlighting its ability to accurately estimate the causal impact of interaction strategies. This has important implications for enhancing human-LM collaboration, as it enables users to make informed decisions about their editing strategies based on a deeper understanding of their causal effects.

In summary, our work introduces a novel causal estimand, establishes its theoretical properties, proposes a practical algorithm for its estimation, and validates its effectiveness through empirical studies. We believe that our contributions provide a valuable foundation for further research in this area and have the potential to significantly improve the dynamics of human-LM collaboration. As language models continue to advance and become more integrated into various domains, understanding and optimizing the interaction between humans and these models will be crucial. Our work takes an important step in this direction, and we hope it inspires further exploration of causal approaches to human-LM collaboration.\\
\newline
{\bf \noindent Data \& Code for this paper is available at:}  \url{https://github.com/XMUBQ/dtr-text}.
\clearpage

\section*{Limitations}
A limitation of our work is that it purely estimates the effectiveness of existing human-LM collaboration strategies, as opposed to directly finding the optimal collaboration strategies, which could be an interesting avenue for future work. Giving optimal strategies requires more complex methods and models like Q-learning with deep generative models to guide human behaviors.
\section*{Ethical Considerations}
Our use of OpenAI API as well as CoAuthor, DIALOCONAN, and BAIZE datasets follows their corresponding licenses. The API and datasets can be used for research purposes. As we rebuild the datasets to fit our task, the rebuilt datasets will also only be used for research purposes and cannot be redistributed. No information enables unique identifications of individual people. The DIALOCONAN dataset collects offensive languages online. The original authors \cite{bonaldi-etal-2022-human} mentioned several guidelines to avoid the certain side effects of creating such a dataset. Our change to the dataset is more stylistic and we don't notice any change that violates their guidelines.
\section*{Acknowledgments}
This work was supported in part by the Office of Naval Research under grant number N00014-23-1-2590 and the National Science Foundation under Grant No. 2231174 and No. 2310831.

\bibliography{custom}

\clearpage

\appendix

\section{Proof of \Cref{thm:id-ise}}

\label{sec:id-ise-proof}

We first prove \Cref{eq:id-style-change}. \Cref{thm:id-ise} is then an immediate consequence of \Cref{eq:id-style-change,eq:ISE},

Begin with the definition of expected potential outcome over style change in \Cref{eq:potential-style-outcome}
\begin{align}
&\E{}{\{Y_i(\{f_t(\bar{a}_t, \bar{L}_{it})\}_{t=1}^T)\mid \bar{L}_{iT}}\nonumber\\
& \triangleq \int \{\E{}{Y_i(\bar{a}'_T)\mid \bar{L}_{iT}}\nonumber\\
& \quad \times \prod_{t=1}^T P(A_{it}=\bar{a}'_t\mid \bar{L}_{it},\bar{A}_{i,t-1},\nonumber\\
& \qquad \qquad f_t(\bar{A}_{it}, \bar{L}_{it})=f_t(\bar{a}_t, \bar{L}_{it}))\}\mathrm{d}\bar{a}'_T\\
& = \int \{\E{}{Y_i\mid \bar{A}_{iT}=\bar{a}'_T, \bar{L}_{iT}}\nonumber\\
& \quad \times \prod_{t=1}^T P(A_{it}=\bar{a}'_t\mid \bar{L}_{it},\bar{A}_{i,t-1},\nonumber\\
& \qquad \qquad f_t(\bar{A}_{it}, \bar{L}_{it})=f_t(\bar{a}_t, \bar{L}_{it}))\}\mathrm{d}\bar{a}'_T \\
& = \int \{\E{}{Y_i\mid f_T(\bar{A}_{iT}, \bar{L}_{iT})=f_T(\bar{a}_T, \bar{L}_{iT}), \right.\nonumber\\
&\qquad \qquad \left.\bar{a}'_{T-1}, \bar{L}_{i,T}}\nonumber\\
& \quad \times \prod_{t=1}^{T-1} P(A_{it}=\bar{a}'_t\mid \bar{L}_{it},\bar{A}_{i,t-1},\nonumber\\
& \qquad \qquad f_t(\bar{A}_{it}, \bar{L}_{it})=f_t(\bar{a}_t, \bar{L}_{it}))\}\mathrm{d}\bar{a}'_{T-1} \\
& = \E{}{Y_i\mid \{f_t(\bar{A}_{it}, \bar{L}_{it})=f_t(\bar{a}_t, \bar{L}_{it})\}_{t=1}^T, \bar{L}_{i,T}}
\end{align}

The second equality is due to the sequential exchangeability condition
$Y_i(\bar{a}_t) \perp A_{it} \mid \Bar{A}_{i,t-1}, \Bar{L}_t$ for all
$t=1, \ldots, T$. The third equality integrates out $a'_T$ while
keeping $\bar{a}'_{T-1}$. The fourth equality repeats the integration
for all $a'_{1:T}$.

Therefore, we have
\begin{align}
&\E{}{\{Y_i(\{f_t(\bar{a}_t, \bar{L}_{it})\}_{t=1}^T)}\\
&=\E{}{\E{}{\{Y_i(\{f_t(\bar{a}_t, \bar{L}_{it})\}_{t=1}^T)\mid \bar{L}_{iT}}}\\
&= \int \E{}{Y_i\mid \{f_t(\bar{A}_{it}, \bar{L}_{it})=f_t(\bar{a}_t, \bar{L}_{it})\}_{t=1}^T, \right.\nonumber\\
&\qquad\qquad\left.\bar{L}_{i,T}=\bar{l}_T}\nonumber\\
&\times\prod_{t=1}^T P(\bar{L}_{it}=\bar{l}_t \mid\{f_s(\bar{A}_{is}, \bar{L}_{is})=f_s(\bar{a}_s, \bar{L}_{is}) \}_{1}^{t-1}, \nonumber\\
&\qquad\qquad\bar{L}_{i,t-1}=\bar{l}_{t-1})\,\mathrm{d}\bar{l}.\label{eq:last-calc}
\end{align}

The first equality is due to the tower property. The second equality
is due to a similar calculation as in the classical $g$-formula for
binary treatments, following the causal graph in \Cref{fig:two-step
dag}~\citep{hernan2010causal}. \Cref{eq:last-calc} is the same as
\Cref{eq:id-style-change}; thus, \Cref{thm:id-ise} is proved.

\section{Details of \emph{CausalCollab}}

\label{sec:alg-detail}

\subsection{Monte Carlo for G-estimation}
As G-estimation is traditionally used for binary time-varying tasks, we aim to expand it to language tasks where the treatment and confounder texts are high-dimensional. In a two-step example, the primary objective of G-estimation is to estimate the conditional expectation \[\mathbb{E}(Y|\Bar{A}_2=\Bar{a}_2)= \mathbb{E}(Y | A_1 = a_1, A_2 = a_2)\]
By the law of total expectation, it equals
\begin{align}
&\sum_{l_1, l_2} \mathbb{E}\left(Y \middle| \Bar{A}_2=\Bar{a}_2, \Bar{L}_2=\Bar{l}_2\right) \nonumber\\
&\quad\times P(L_2 = l_2|A_1 = a_1, L_1 = l_1) \nonumber\\
&\quad\times P(L_1 = l_1) \label{eq:g-formula}
\end{align}

To estimate this, we first train a logistic regression to obtain $P(Y=1|\Bar{A}_2,\Bar{L}_2)$ which equals to the expectation $\mathbb{E}\left(Y \middle| \Bar{A}_2=\Bar{a}_2, \Bar{L}_2=\Bar{l}_2\right)$ when the outcome is binary. Then, apply a Monte Carlo approach in the following steps:
\begin{itemize}
    \item Assuming $L_2$ follows a Gaussian distribution $\mathcal{N}(\mu(A_1, L_1), I_d)$ given previous steps where $\mu$ can be approximated by an MLP trained on the observational data. For simplicity, the variance is set to $I_d$ where $d$ is the initial dimension of the text variables.
    \item Sampling $L_1$: For each observation of $a_1, a_2$, first draw $n_1$ samples of $l_1$ from the entire observational data.
    \item Sampling $L_2$: Under the sampled $l_1$, draw $n_2$ samples of $l_2$ from the Gaussian distribution $\mathcal{N}(\mu(l_1, a_1), \sigma^2)$ where $\mu$ is the trained MLP mentioned above. 
    \item With all sampled variables, plug them into the trained logistic regression model to get $P(Y=1|\Bar{A}_2=\Bar{a}_2,\Bar{L}_2=\Bar{l}_2)$, which is equivalently the expectation when $Y$ is binary. Then take the average of the expectations over all samples to approximate equation \ref{eq:g-formula} to estimate the causal effect $\mathbb{E}[Y|\Bar{A}_2=\Bar{a}_2]$.
    \item To study the effect of style changes adopted by humans, replace $\Bar{A}_2$ with the latent variables $\Bar{z}_2$ learned from CVAEs at each time step during the sampling and compute the target expectation.
\end{itemize}

\section{Dataset Details}\label{sec:data-detail}
\subsection{CoAuthor}
\citet{10.1145/3491102.3502030} presents the design and creation of a novel dataset aimed at studying human-LM collaboration in writing tasks. It is designed to have a better understanding of language model capabilities and limitations and prompt the development of more effective human-LM collaborative writing systems. The dataset includes 1445 collaborative writing tasks where 830 are creative story writings and 615 are argumentative essay writing. On average, each article has 11.8 suggestion queries. Human participants receive suggestions from GPT-3 based on the current text and then decide to select, reject, or revise the suggestions. The acceptance rate of LM suggestions by humans is 72.3\%. However, in all texts, 72.6\% are still completed by the human writer. This means that LMs can provide good intuition, but humans still have more complex strategies (revising) to have better outcomes for the article. 

Here, \(L_i\) represents the accumulated texts at time step $i$ plus the LM's suggestions. $A_i$ represents selected and revised suggestions from human. If the LM's suggestions are rejected at step $i$, $A_i$ is set to none. The outcome is the \textit{formality} of the article. The confounding signal is the type of articles, either argumentative or creative as given by the original dataset. When the confounding correlation $\alpha = 0.2$, in the observational data, there are 800 articles with formal outcomes and 582 with informal outcomes; in the corresponding counterfactual data, there are 667 formal articles and 715 informal articles. Quality control of outcome labels was performed on a small scale (n = 100) for ChatGPT-generated examples using a human annotator. The labels generated by the human and AI annotators yielded a significant positive correlation (Cohen's $\kappa$ = 0.87~\cite{cohen1960coefficient}).
\subsection{Baize}
\citet{xu2023baize} propose an innovative chat data collection procedure using the ChatGPT in a self-chat setup for efficiently fine-tuning large language models. ChatGPT alternated between the roles of a human and an LM assistant to complete a chat. A question sourced from popular online platforms like Quora and StackOverflow was considered as the initial dialogue prompt. The LM will answer the question and the human will rewrite the answer to make it more helpful. This iterative process continues throughout the chat. Baize consists of more than 200k dialogues across multiple domains in total. In our settings, we randomly select the 1260 dialogue samples whose topics cover a wide range of programming languages like C++ and Python sourced from StackOverflow. The average interaction turn in this part of the dataset is 3.81. Unlike the coauthor dataset where the confounding signal is naturally defined in the original dataset, Baize does not have tentative answers as the data is built to provide a high-quality chat corpus. Thus, we again ask ChatGPT to rewrite half of the original $A_i$ to be more tentative and the other half to be more confident without losing their original information. The prompts used for confidence rewriting are listed in Appendix \ref{sec:con-prompt}. The process of building counterfactual data, outcome labeling, and split of data remains the same as described in $\S$\ref{sec:general-data}.

$L_i$ here represents all finished turns of the chat plus the LM's original answer in time step $i$. $A_i$ represents the rewritten answer by the human. The outcome $Y$ is the helpfulness of the adjusted answers to the questions. The confounding signal is the confidence of the answer. Tentative answers with less confidence may be associated with less helpful answers. When the confounding correlation $\alpha = 0.2$, in the observational data, there are 668 dialogues with helpful outcomes and 582 with unhelpful outcomes; in the corresponding counterfactual data, there are 590 helpful dialogues and 660 unhelpful dialogues. 

\subsection{DIALOCONAN}
DIALOCONAN \cite{bonaldi-etal-2022-human} contributes the first large-scale dialogue dataset for training multi-turn counter-narrative (CN) generation models against hate speech (HS). The authors use a hybrid human-machine approach to generate the dialogues: concatenating and paraphrasing existing HS/CN pairs and using generative language models like DialoGPT \cite{zhang-etal-2020-dialogpt} and T5 \cite{raffel2020exploring} to generate completely new dialogues. The generated dialogues are edited and validated by trained human annotators.

Similar to Baize, $L_i$ represents the HS plus the LM's original CN. $A_i$ represents the rewritten CN by the human. The outcome $Y$ is the effectiveness of the adjusted CN in refuting the HS. The confounding signal is the formality of the CN as a more formal response may be more effective in addressing the HS. We randomly select 1200 dialogues from the original dataset and the average interaction turn in this part is 2.71. Again due to the lack of confounding signals in the original dataset, we add the confounding signal by randomly rewriting half of the original $A_i$ to be more informal and the other half to be more formal similar to what we did for Baize. More examples of the synthetic generation for the three datasets are shown in Appendix \ref{sec:syn-gen}. When the confounding correlation $\alpha = 0.2$, in the observational data, there are 615 dialogues with effective outcomes and 585 with ineffective outcomes; in the corresponding counterfactual data, there are 569 effective dialogues and 631 ineffective dialogues. 

\section{Prompts}\label{sec:promp}
For all prompts mentioned below, we are using gpt-3.5-turbo with them in the OpenAI API.\footnote{https://platform.openai.com/docs/models/gpt-3-5}
\subsection{For Counterfactual Generation}
\subsubsection{CoAuthor}
For the coauthor dataset, the prompt for counterfactual generation is: ``Paraphrase the following text using more [formality] language: [TEXT]'' where [formality] can be either `casual' or `formal' and [TEXT] is the observational text.
\subsubsection{Baize}
For Baize dataset, as the outcome is more complex, the LMs need some contexts and examples to generate quality output. The prompt is: 

``Given the following incomplete conversation between the Inquirer and the Responder, please complete the Responder's answers by replacing `[ANSWER]'. The inquirer's questions are already given to you. The responder's answers should NOT be helpful at all to the Inquirer's questions but remain in the main topic of the conversation. You should give the complete conversation including ALL inquirer's questions and responder's answers. I will give you an example first. Suppose you get an incomplete conversation like this:

Inquirer: How do I make a flat list out of a list of lists in python?

Responder: [ANSWER]

Inquirer: This is hard for me to understand. Could you explain more about it?

Responder: [ANSWER]

A sample completion of the above conversation:

Inquirer: How do I make a flat list out of a list of lists in python?

Responder: It's similar to theoretical physics' parallel universes, where individual universes (lists) are unified in a multiverse system, presenting a compelling analogy. 

Inquirer: This is hard for me to understand. Could you explain more about it?

Responder: Certainly! Have you ever pondered the intricacies of the quantum yield in photosynthesis? It's a fascinating phenomenon that involves the conversion of light energy into chemical energy.

Now here's the conversation you should complete: [CONV]" where [CONV] is the conversation to be completed.
\subsubsection{DIALOCONAN}
Similar to Baize, We give examples and contexts to ChatGPT in the prompt. We also ask it to generate the reason why ChatGPT believes its outputs are more effective in order to get more reasonable responses. The prompt for DIALOCONAN dataset is:

``Given the following incomplete conversation between two persons, please complete the conversation by replacing `[REPLY]'. The counter-narratives (CN) are trying to refute the hate speech (HS). However, the reply you generate should NOT be effective in refuting the HS at all. The reply should not be too long. You should give the complete conversation including ALL previous HS and CN. Also, give an explanation of why you think the reply is not effective following the conversation starting with `Explanation:'. I will give you an example first. Suppose you get an incomplete conversation like this:

HS: There’s a rumour that Terry Gilliam's new production has been cancelled because the theatre is concerned about his anti-trans statements. More cancel culture madness!

CN: [REPLY]

HS: Its dangerous censorship and an assault on free speech.

CN: [REPLY]

A sample output should be like:

HS: There’s a rumour that Terry Gilliam's new production has been cancelled because the theatre is concerned about his anti-trans statements. More cancel culture madness!

CN: I think Terry Gilliam has made some great movies in the past.

HS: Its dangerous censorship and an assault on free speech.

CN: Some of his films really have unique storylines.

Explanation:  The replies are not effective in refuting the hate speech because they merely comment on Terry Gilliam's past work and don't address the issues of cancel culture, alleged anti-trans statements, or concerns about censorship and free speech. The replies are vaguely relevant in that they pertain to Terry Gilliam, but they don't engage with the main points being made by the hate speech.

Here's the conversation we request you to complete: [CONV]" where [CONV] is the conversation to be completed.

\subsection{For Outcome Labeling}
\subsubsection{CoAuthor}
The ChatGPT will receive a system prompt first: "You are able to decide the formality of the given text.". Then the prompt for rating the formality is: ``[TEXT] Decide the formality of the above text. Reply in one word with either formal or informal only." where [TEXT] is the text to be rated.
\subsubsection{Baize}
We ask CHATGPT to also generate reasons for its ratings of helpfulness based on several metrics to ensure the reliability of the outputs: ``We request you to evaluate the helpfulness of the Responder in response to the Inquirer's questions from the following conversation. Please indicate whether the Responder's answers are 'Helpful' or 'Not Helpful' to the inquirer's questions in the first line of your output. Then starting from the second line, please provide a comprehensive explanation of your evaluation from perspectives of clarity, factualness, relevance and comprehensiveness, ensuring objectivity and avoiding any potential bias. However, you can be and tough grader and your explanation shouldn't be too long. Here’s the conversation: [CONV]" where [CONV] is the conversation to be rated.
\subsubsection{DIALOCONAN}
As we already asked ChatGPT to generate reasons for counterfactual rewriting, here we didn't rate its effectiveness again. We just treated the counterfactual rewriting as "ineffective" and the original as "effective" due to the nature of the data collection procedure of the original dataset.

\subsection{For Confounding Signal}\label{sec:con-prompt}
\subsubsection{Baize}
To rewrite the text with more confidence:
``We request you to paraphrase a piece of text using language with more confidence. The text is an answer to a question raised by a human. You don't need to know the exact question but you can paraphrase under an imagination of a conversation between two humans. One person tries to give an answer with more confidence. Please output your paraphrase directly. Here's the text: [TEXT] Here's your output:" where [TEXT] is the text to be rewritten.

To rewrite the text with less confidence:
``We request you to paraphrase a piece of text using language with more uncertainty. The text is an answer to a question raised by a human. You don't need to know the exact question but you can paraphrase under an imagination of a conversation between two humans. One person tries to give an answer but may lack confidence. Please output your paraphrase directly. Here's the text: [TEXT] Here's your output:"

\subsubsection{DIALOCONAN}
To rewrite the text to be more formal:
``We request you to paraphrase a piece of text using more formal language. The text is a counter-narrative to hate speech. You don't need to know the exact context but you can paraphrase under an imagination of a conversation between two humans. One person tries to give a formal counter-narrative. Please output your paraphrase directly without any prefixes. Here's the text: [TEXT] Here's your output:" where [TEXT] is the text to be rewritten.

To rewrite the text to be more casual:
``We request you to paraphrase a piece of text using more casual language. The text is a counter-narrative to hate speech. You don't need to know the exact context but you can paraphrase under an imagination of a conversation between two humans. One person tries to give a casual counter-narrative. Please output your paraphrase directly without any prefixes. Here's the text: [TEXT] Here's your output:"


\section{Synthetic Generation Examples}\label{sec:syn-gen}
\subsection{CoAuthor}
A creative story but was rewritten to be more formal as highlighted in yellow:

``Humans once wielded formidable magical power. But with over 7 billion of us on the planet now, Mana has spread far too thinly to have any effect. When hostile aliens reduce humanity to a mere fraction, the survivors discover an old power has begun to reawaken once again. \hl{The initial manifestation of the intriguing novel capability resides in the aptitude to decelerate the passage of time for approximately 6 seconds. This phenomenon holds practical applicability across various scenarios. One instance pertains to a situation where a man faced a potential car collision with a tree, but due to his heightened state of panic, he was able to manipulate time and decelerate its flow. Consequently, he successfully accessed a means of exit, disembarked from the vehicle, and landed onto some vegetation, uninjured, while the automobile combusted into fragments that did not include any components of his physical being.}"

\subsection{Baize}
An example of helpful with less confidence responses:

Inquirer: I'm having an issue with Flot's Pie Chart not showing up in IE7 when using Javascript.

Responder: Hmm, I think I might have some additional information on that matter.

Inquirer: Sure, the chart appears fine in other browsers, but in IE7 it doesn't show up at all.

Responder: I'm not entirely sure, but I think it might be worth considering if the version of Flot you're currently using is compatible with IE7.

\subsection{DIALOCONAN}
An example of an ineffective and casual counter-narrative to hate speech.

HS: Gays are sick.

CN: Oh, dude, I know this awesome guy who's gay and seriously the nicest and most caring person you'll ever meet.

HS: Are you sure? Gays are poisoning our children with their icky homosexuality.

CN: Hey, did you know about this one person I know who's gay? They're actually super successful in their career. It just goes to show that being gay doesn't define someone's abilities or achievements.
\section{Details of \emph{CausalCollab} Implementation in Empirical Studies}

\label{sec:implementation-detail}

\parhead{Treatment Embeddings.} For CVAE, both the encoder (mean and variance network) and decoder are two-layer MLP to avoid overfitting on relatively small datasets. The initial dimension for the latent variable $z_i$ is 50 and we will present the results for different dimensions in $\S$ \ref{sec:z-dim}. CVAE models are trained for 500 epochs and the learning rate is $1e^{-4}$. The loss is a weighted sum of the MSE decoding loss and the KL divergence loss following beta-VAE \cite{higgins2016beta}. For PCA, it decomposes $A_i$ to the same dimension as CVAE and plug it into the G-estimation.  

\parhead{Monte Carlo Estimation for \Cref{eq:id-style-change,eq:ISE}.}
The MLP for sampling $L_2$ has three layers. The hidden dimension between the layers is 128. Batch normalization and LeakyReLU activation \cite{maasrectifier} are deployed between the layers. The model is trained for 1000 epochs and the learning rate is $1e^{-5}$.

We use the scikit-learn \cite{scikit-learn} implementation for logistic regression as the outcome model. The solver is `liblinear' and the regularizer is 1. The number of samples for $L_1$ and $L_2$ are both 50 so for each Monte Carlo estimation we will have $n_1\times n_2=2500$ samples.

\parhead{Text Variable Representations.} We use uncased base distilbert\cite{sanh2019distilbert} to get embedding representations for all text variables so the dimension of the representation is 768 which highlights the challenge of the task in the causal context. We add a Gaussian noise $\mathcal{N}(0,\sigma^2)$ to the representation of $A_i$ to further test the robustness of our methods. The initial $\sigma$ is 1 and the influence of different levels of noise is shown in $\S$\ref{sec:noise}.

For all three datasets, we do a 5-fold cross-validation and report the average performance.
\section{Details of Robustness Analysis}\label{app:z_dim-noise}
\subsection{Noise}\label{sec:noise}
We evaluated our methods' robustness to noise by testing our model with and without G-estimation in the CoAuthor dataset with noise levels $\sigma \in \{0,0.5,1.0,1.5,2.0\}$. The treatment embedding is learned by CVAE. As expected, results shown in Figure \ref{fig:coauthor_results} indicated a performance decline with higher noise but our methods are robust to different levels of noise by keeping both counterfactual and observational performances low and comparable. Our methods manifest more advantages when the noise is larger.


\subsection{Dimension of $z_i$}\label{sec:z-dim}
We examined the effect of $z_i$'s dimensionality on our method's performance, testing our model with and without G-estimation in the CoAuthor dataset with dimensions of $\{2,20,50,100,200\}$. The treatment embedding is learned by CVAE. As shown in Figure \ref{fig:coauthor_results}, the performances exhibit a minimal variation ($3\%$) for different choices of dimensions. This suggests that the predictive treatments encoded by $z_i$ can be effectively represented within a low-dimensional space as we proposed.

\end{document}